\documentclass[11pt]{article}
\usepackage{coling2020}
\usepackage{times}
\usepackage{url}
\usepackage{latexsym}
\usepackage{cleveref}
\usepackage{multirow}
\usepackage{booktabs}
\usepackage{makecell}
\usepackage[usestackEOL]{stackengine}
\usepackage{graphicx}
\usepackage{booktabs}
\usepackage{longtable}
\usepackage{lipsum} 
\usepackage[toc,page]{appendix}
\usepackage{afterpage}

\colingfinalcopy 

\title{\textsc{KinNews} and \textsc{KirNews}: Benchmarking Cross-Lingual Text Classification for Kinyarwanda and Kirundi}

\author{
Rubungo Andre Niyongabo$^{1,2}$ \and Hong Qu$^{1,}$\thanks{\hspace{0.13cm}Corresponding author} \and Julia Kreutzer$^{2}$ \and  Li Huang$^{1}$  \\
 $^{1}$School of Computer Science and Engineering, Computational Intelligence Lab,\\ University of Electronic Science and Technology of China, China\\
 $^{2}$Masakhane, Africa\\
 {\tt niyongabor.andre@std.uestc.edu.cn, hongqu@uestc.edu.cn,}\\
 {\tt juliakreutzer.jul@gmail.com, lihuang@std.uestc.edu.cn}
}

\date{}

\begin{document}

\strutlongstacks{T}
\maketitle
\begin{abstract}
  Recent progress in text classification has been focused on high-resource languages such as English and Chinese. For low-resource languages, amongst them most African languages, the lack of well-annotated data and effective preprocessing, is hindering the progress and the transfer of successful methods.  In this paper, we introduce two news datasets (\textsc{KinNews} and \textsc{KirNews}) for multi-class classification of \emph{news} articles in \emph{Kin}yarwanda and \emph{Kir}undi, two low-resource African languages. The two languages are mutually intelligible, but while Kinyarwanda has been studied in Natural Language Processing (NLP) to some extent, this work constitutes the first study on Kirundi. Along with the datasets, we provide statistics, guidelines for preprocessing, and monolingual and cross-lingual baseline models. Our experiments show that training embeddings on the relatively higher-resourced Kinyarwanda yields successful cross-lingual transfer to Kirundi.  In addition, the design of the created datasets allows for a wider use in NLP beyond text classification in future studies, such as representation learning, cross-lingual learning with more distant languages, or as base for new annotations for tasks such as parsing, POS tagging, and NER. The datasets, stopwords, and pre-trained embeddings are publicly available at \url{https://github.com/Andrews2017/KINNEWS-and-KIRNEWS-Corpus}. 
\end{abstract}

\section{Introduction}
\label{intro}

%
%
\blfootnote{
    %
    %
    %
    %
    %
    %
     \hspace{-0.65cm}  
     This work is licensed under a Creative Commons 
     Attribution 4.0 International License.
     License details:
     \url{http://creativecommons.org/licenses/by/4.0/}.
}

The availability of large monolingual and labeled corpora, paired with innovations in neural text processing have led to a rapid improvement of the quality of text classification over the last years.\footnote{See e.g. \url{https://nlpprogress.com/english/text_classification.html} or \url{https://paperswithcode.com/task/text-classification}}  
However, the effectiveness of deep-learning-based text classification models depends on the amount 
of monolingual and labeled data. Low-resource languages are traditionally left behind because of the few available prepared resources for these languages to extract the data from~\cite{joshi-etal-2020-state}. However, nowadays, the increase in internet use in many African developing countries has made access to information easier. This in turn has strengthened the news agencies of those countries to cover many stories in their native languages. For example, \textit{BBC News} now provides online news in Arabic, Amharic, Hausa, Kiswahili, Somali, Oromo, Igbo, Nigerian Pidgin, Tigrigna, Kinyarwanda and Kirundi.\footnote{\url{https://www.bbc.co.uk/ws/languages}} This development makes news the most reliable source of data for low-resource languages. We explore this opportunity for the example of Kinyarwanda and Kirundi, two African low-resource Bantu languages, and build news classification benchmarks from online news articles. This has the goal to enable NLP researchers to include Kinyarwanda and Kirundi in the evaluation of novel text classification approaches, and diversify the current NLP landscape.

\textit{Kinyarwanda} is one of the official languages of Rwanda\footnote{\url{https://en.wikipedia.org/wiki/Kinyarwanda}} and belongs to the Niger-Congo language family. According to \textit{The New Times},\footnote{\url{https://www.newtimes.co.rw/section/read/24728}} it is spoken by approximately 30 million people from four different African countries: Rwanda, Uganda, DR Congo, and Tanzania. \textit{Kirundi} is an official language of Burundi, a neighboring country of Rwanda, and it is spoken by at least 9 million people.\footnote{\url{https://en.wikipedia.org/wiki/Kirundi}} Kinyarwanda and Kirundi are mutually intelligible. 
Table~\ref{kin-kir-sim-ex} shows two example sentences to illustrate the similarity of the languages. The two languages are part of the wider dialect continuum known as Rwanda-Rundi.\footnote{\url{https://en.wikipedia.org/wiki/Rwanda-Rundi}} In this family, there are other four indigenous low-resource languages: \textit{Shubi},\footnote{\url{https://en.wikipedia.org/wiki/Shubi_language}} \textit{Hangaza},\footnote{\url{https://en.wikipedia.org/wiki/Hangaza_language}} \textit{Ha},\footnote{\url{https://en.wikipedia.org/wiki/Ha_language}} and \textit{Vinza}.\footnote{\url{https://en.wikipedia.org/wiki/Vinza_language}} Among these four languages, \textit{Ha} is also mutually intelligible for Kirundi and Kinyarwanda speakers, while other three are partially mutually intelligible. Developing NLP models for these languages is the goal for future work, because we could not retrieve any written news data for them.

\begin{table}[ht]
\begin{center}
\resizebox{\columnwidth}{!}{%
\begin{tabular}{ll}
\toprule \bf Language & \bf Sentence \\ \midrule
Kirundi & \textit{Turafise \textbf{ubwoba} \textbf{ko} \textbf{inyuma} \textbf{y’ikiza} ca \textbf{coronavirus} bazohava bakena cane \textbf{kuruta} \textbf{mbere.}}\\
Kinyarwanda & \textit{Dufite \textbf{ubwoba} \textbf{ko} \textbf{inyuma} \textbf{y'ikiza} cya \textbf{coronavirus} bashobora gukena cyane \textbf{kuruta mbere.}}\\ 
English & \textit{We fear that after the coronavirus epidemic, they may become poor than before.}\\\midrule
Kinyarwanda & \textit{\textbf{Ashushanya inyama} n’ibintu \textbf{bikozwe mu masashe n’impapuro.}}\\
Kirundi & \textit{\textbf{Ashushanya inyama} n’ivyamwa \textbf{bikozwe mu masashe n’impapuro.}}\\
English & \textit{He draws meat and things made of sacks and paper.}\\
\bottomrule
\end{tabular}
}
\end{center}
\caption{\label{kin-kir-sim-ex} Two examples of news titles from our datasets that show the similarity level between Kinyarwanda and Kirundi. Same words in both languages for each sentence are shown in bold.}
\end{table}

\newcite{joshi-etal-2020-state} classify the state of NLP for both Kinyarwanda and Kirundi as \textit{``Scraping-By''}, which means that they have been mostly excluded from previous NLP research, and require the creation of dedicated resources for future inclusion in NLP research. To this aim, we introduce two datasets \textsc{\textsc{KinNews}} and \textsc{KirNews} for multi-class text classification in this paper. They consist of the news articles written in Kinyarwanda and Kirundi collected from local news websites and newspapers. \textsc{KinNews} samples are annotated using fourteen classes while that of \textsc{KirNews} are annotated using twelve classes based on the agreement of the two annotators for each dataset. We describe a data cleaning pipeline, and we introduce the first ever stopword list for each language for preprocessing purposes. We present word embedding techniques for these two low-resource languages, and evaluate various classic and neural machine learning models. Together with the data, these baselines and preprocessing tools are made publicly available as benchmarks for future studies. In addition, pre-trained embeddings are published to facilitate studies for other NLP tasks on Kinyarwanda and Kirundi.

In the following, we will first discuss previous work on Kinyarwanda and Kirundi and low-resource African languages in general in Section~\ref{sec:related}, and then describe the dataset creation in Section~\ref{sec:data}. We then present a range of experiments for text classification on the collected data in Section~\ref{sec:experiments}, concluding with an outlook to future work in Section~\ref{sec:conclusion}. 

\section{Related Work}\label{sec:related}
\newcite{joshi-etal-2020-state} introduce a taxonomy of languages in NLP that expresses to which degree they have been subject to NLP research until 2020. Many of the roughly 2000 African languages fall under the categories \textit{``Scraping-By''} or \textit{``Left-Behind''}, which means that they have been systematically understudied or ignored in NLP research. As \newcite{joshi-etal-2020-state} discuss, those languages require dedicated, often manual effort to enable NLP research, since even monolingual digital data does not exist or is hard to find on the web. 
In recent years, NLP research on African languages has been increasing thanks to new datasets being published, such as the \textit{JW300} corpus \cite{agic-vulic-2019-jw300}, which provides parallel data for 300 languages and facilitated machine translation research for many low-resource languages~\cite{TiedemannThottingal:EAMT2020}, amongst them many African languages \cite{orife2020,masakhane} which have not been subject to machine translation before.

Beyond the multilingual \textit{JW300} corpus, there have been a few works have been done for creating new datasets for individual African languages. For example,  \newcite{emezue-dossou-2020-ffr} introduced the FFR project for creating a corpus of Fon-French (FFR) parallel sentences. Closer to our work, \newcite{marivate-etal-2020-investigating} created news classification benchmarks for Setswana and Sepedi, two Bantu languages of South Africa. Different to our work, their corpus is limited to headlines, while we provide headlines and the full articles. The size of our dataset is also several magnitudes larger since we include data from more sources and spent large efforts on expanding an initial set of news sources.

While there is practically no NLP research on Kirundi, there are a few recent studies on Kinyarwanda for the tasks of Morphological Analysis \cite{muhirwe2009morphological}, Part-of-Speech (POS) tagging \cite{garrette2013learning,garrette2013real,fang-cohn-2016-learning,duong2014can,cardenas-etal-2019-grounded}, Parsing \cite{sun2014parsing,mielens2015parse}, Automated Speech Recognition \cite{dalmia2018domain}, Language Modeling \cite{andreas-2020-good}, and Name Entity Recognition \cite{rijhwani-etal-2020-soft}. Most of these works are largely based on a single Kinyarwanda dataset created by \newcite{garrette2013learning}. This dataset contains transcripts of testimonies by survivors of the Rwandan genocide, was provided by the Kigali Genocide Memorial Center, and contains 90 annotated sentences with fourteen distinct POS tags. However, it is not suitable for text classification, and to the best of our knowledge, there are no  publicly available datasets for Kinyarwanda and Kirundi text classification, which is the gap that this paper is addressing. These works have also focused on either word alignment or monolingual approaches, and did not explore a cross-lingual approach.

We hope that the publication of our benchmarks will inspire the creation of similar datasets. As a result, this would allow the inclusion of more African low-resource languages in cross-lingual studies and benchmarks such as \textit{XTREME} \cite{hu2020xtreme}, a multi-task benchmark for the evaluation of cross-lingual generalization of multilingual
representations across 40 languages, which already include higher-resourced African languages like Afrikaans and Swahili. For past efforts of multi-lingual studies like \textit{XTREME}, one guiding factor for language selection has been the size of the Wikipedia in the respective languages. The number of Wikipedia articles in local languages is often interpreted as a measure for digital maturity and a pragmatic estimator for the success of un- or self-supervised NLP methods, but this ignores societal and human factors that (cyclically) influence the activity of Wikipedia editor communities. In this work, we want to showcase the impact of manual collection of data sources beyond Wikipedia. The number of news articles that we could retrieve for Kirundi and Kinyarwanda exceeds the number of available Wikipedia articles by far (616 and 1828 Wikipedia articles respectively).\footnote{\url{https://meta.wikimedia.org/wiki/List_of_Wikipedias} as of June 29, 2020.}

\section{Dataset Creation}\label{sec:data}
In this section, we first describe the process for data collection, then the annotation, and finally our data cleaning pipeline for \textsc{\textsc{KinNews}} and \textsc{KirNews}. 
In general, the copyright for the published news articles still remains with the original authors or publishers. Our work can be seen as an additional pre-processing and modeling pipeline, and annotation layer on top.

\subsection{Collection Process}
\paragraph{\textsc{KinNews}}
\textsc{KinNews} is collected from fifteen news websites and five newspapers from Rwanda. An initial seed of news sources was retrieved from two websites which list newspapers from Rwanda.\footnote{\url{https://www.w3newspapers.com/rwanda/} and \url{http://www.abyznewslinks.com/rwand.htm}} These lists also include Rwandan news sources that publish their news in other languages such as French and English, so we select those which publish in Kinyarwanda only. Additionally, we expand the initial seed through Google Search by searching for manually selected Kinyarwanda key words and phrases such as \textit{``Iterambere ry'umugore mu Rwanda''} (\textit{``Women's development in Rwanda''}) to list all news sources that have published the searched key phrase or related news. These were used to expand the list of news sources that publish their news in Kinyarwanda. 

\paragraph{\textsc{KirNews}}
We used the same process for \textsc{KirNews}. \textsc{KirNews} was collected from eight news sources in total. However, it was more challenging, since most of the news sources that were listed on the overview websites\footnote{\url{https://www.w3newspapers.com/burundi/} and \url{http://www.abyznewslinks.com/burun.htm}} publish their news either in French or English. Only one news website was found that publishes in Kirundi. Thus, to solve this problem, the same seed expansion technique as described above was used to find three more news websites and four newspapers that publish in Kirundi, which was very time-consuming. We hope that this kind of seed expansion can in future be automated with the help of NLP technology for Kirundi.

\paragraph{Document Structure} Each data sample for both \textsc{KinNews} and \textsc{KirNews} consists of the news headline and the article's content. We separate the title from the content to make the annotation process easier, since the annotator may sometimes simply annotate the news based on its title without reading the whole article. In addition, the original source URLs are recorded with the extracted article, such that meta-information about dates or authors, or multi-modal content such as embedded images and captions can be retrieved if needed. Future NLP studies may also exploit this structure, e.g. for headline prediction or automatic summarization.

\subsection{Annotation Process}
The news we collected were initially given different categories by the publishers. The articles in \textsc{KinNews} and \textsc{KirNews} were categorized in 48 and 26 different categories, respectively. However, many of these categories were related and to reduce the noisy samples, annotators agreed in grouping the related categories into one category and finally resulted into fourteen categories for \textsc{KinNews} and twelve categories for \textsc{KirNews} in total (details in Appendix~\ref{appendix:table}).  

In both datasets, each category was assigned with its own numerical labels (\texttt{label}) that range from 1 to 14 and English labels (\texttt{en\_label}) to help those who do not understand Kinyarwanda and Kirundi to know what the article is related to. Moreover, Kinyarwanda labels (\texttt{kin\_label}) for \textsc{KinNews} and Kirundi labels (\texttt{kir\_label}) for \textsc{KirNews} were also provided.

Then, based on these agreed categories, two annotators for each dataset who are all linguistic graduates and native speakers of each language, attentively revised each news article and annotate it based on its title and content. If they encountered one article that they would hesitate on its category, they annotated it as \textit{neutral} and it receives a numerical label of 0, to be later removed from the final dataset to focus on clearly classifiable data.

\subsection{Dataset Cleaning}
For each language, we provide a cleaned version and a raw version. The cleaning is done in two stages: (1) removal of special characters, and (2) stopword removal. Nowadays, low-resource languages lack of language processing tools and resources \cite{baumann2014using,muis2018low}. Kinyarwanda and Kirundi are also among those languages which do not have any language-specific processing tools (tokenizers, lemmatizers, stemmers, stopword filters, normalizers etc.). 

\paragraph{Special Characters Removal}
The retrieved documents from the internet are often too noisy. To obtain high-quality and cleaned datasets, we remove the following non-alphanumerical characters $\textit{;.?/\textbackslash\textbar\_\@\#\$\%\textendash\textless\textgreater()[]\{\}\&*˜‘+-=ˆ\textbackslash{n}\textbackslash{r}\textbackslash{t}}$ and URLs from the text. Note that removing punctuation might lead to losing sentence boundary information within the article. However, this does not hurt the performance of our models, because they were trained based on word-based features within each article. And since the raw data is provided, punctuation might be restored for other applications, for example for developing language-specific tokenizers.

\begin{figure}[ht]
\centering
\includegraphics[width=\linewidth]{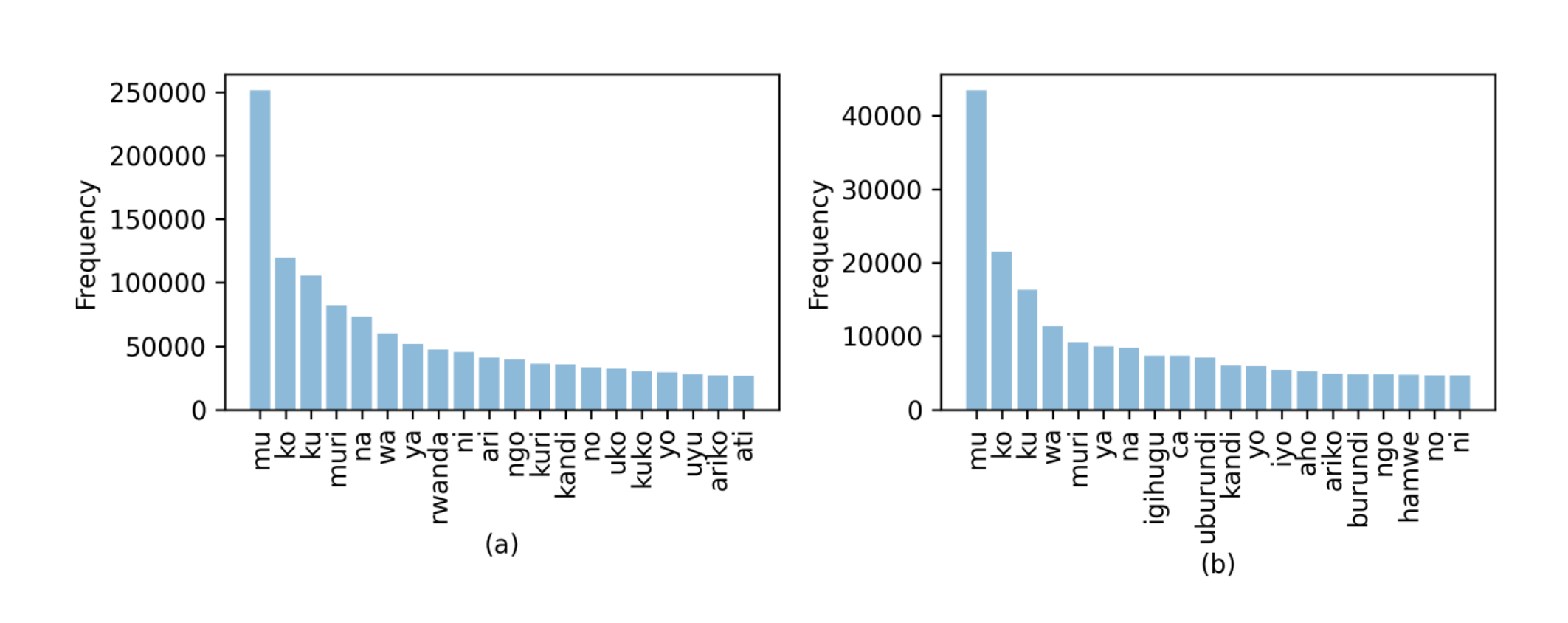}
\caption{\label{fig1} Top 20 tokens in (a) Kinyarwanda and (b) Kirundi data.} 
\end{figure}

\paragraph{Stopwords}
Since stopwords play an important role in semantic text preprocessing, we create first stopword lists for both Kinyarwanda and Kirundi languages. Using additional data from the Kinyarwanda Bible\footnote{\url{https://bibiliya.com}} and based on the sufficient knowledge the annotators have on both languages, we found that the words with two letters such as \textit{``mu'':``in''} and \textit{``ku'':``on''/``at''}, words with three letters such as \textit{``uyu'':``this''} and \textit{``iyo'':``that''}, and words with four letters such as \textit{``muri'':``in''} have high frequency, see Figure~\ref{fig1}, but do not carry a significant role in training text classification models. Thus, these words and other similar words were used to create a list of 80 stopwords for Kinyarwanda and 59 stopwords for Kirundi, listed in Table~\ref{stopword}. The words found in the stopword lists were then removed from the respective cleaned news datasets.

\begin{table}[ht]
\begin{center}
\begin{tabular}{ll}
\toprule \bf Kinyarwanda Stopwords & \bf Kirundi Stopwords \\ \midrule
aba, abo, aha, aho, ari, ati, aya, ayo, ba, baba,
&aba, abo aho, ari,ata, ati, ayo, ba,\\
 babo, bari, be, bo, bose, bw, bwa, bwo, by, bya,
&bari, bo, bose, bw, bwa, bwo, ca, cane,\\ 
 byo, cy, cya, cyo, hafi, ibi, ibyo, icyo, iki, imwe,
&co, de, ico, iryo, ivyo, iyo, izo,ko,\\
 iri, iyi, iyo, izi, izo, ka, ko, ku, kuri, kuva,
&ku, kuri, kuva, kw, maze, mu, muri, mw,\\
kwa, maze, mu, muri, na, naho, nawe, ngo, ni,
&na, naho, nayo, ngo, ni, nk, no, rero,\\
niba, nk, nka, no, nta, nuko, rero, rw, rwa, rwo,
&rw, ry, rya, ubu, uko, uri, uwo, uyu,\\
ry, rya, ubu, ubwo, uko, undi, uri, uwo, uyu, wa,
&vy, vya, vyo, wa, wo, ya, yari, yo,\\
wari, we, wo, ya, yabo, yari, ye, yo, yose, za, zo  
&yose, za, zo\\
\bottomrule
\end{tabular}
\end{center}
\caption{\label{stopword}Stopword list for Kinyarwanda and Kirundi. }
\end{table}

\subsection{Dataset Statistics}
The datasets contain a total of 21,268 and 4,612 news articles which are distributed across 14 and 12 categories for \textsc{KinNews} and \textsc{KirNews}, respectively. As shown in Table~\ref{tab_lbl_stat} \textit{Politics} related articles are the majority in both datasets, while \textit{education} and \textit{history} related articles are the minority in \textsc{KinNews} and \textsc{KirNews}, respectively.

\begin{table}[ht]
\begin{center}
\begin{tabular}{lrllr}
\toprule
\multicolumn{2}{c}{\bf \textsc{KinNews}} & &\multicolumn{2}{c}{\bf \textsc{KirNews}} \\ 
\texttt{en\_label} & \# Articles & & \texttt{en\_label} & \# Articles\\ \hline
Politics & 4,908 & & Politics & 1,642 \\
Economy & 4,418 & & Sport & 1,400 \\
Entertainment & 3,229 & & Economy & 388 \\
Sport & 3,104 & & Relationship & 328 \\
Health & 1,914 & & Health & 264 \\
Relationship & 800 &&  Education & 206 \\
Religion & 782 & & Entertainment & 136 \\
Technology & 439 & & Technology & 70 \\
Culture & 436 & & Environment & 58 \\ 
Tourism & 378 & & Culture & 52 \\ 
History  & 256 & & Religion & 48 \\
Environment  & 213 & & History & 20 \\
Fashion  & 205 & & &\\
Education  & 186 & & &\\
\midrule
\textbf{Total} & 21,268 & & \textbf{Total} & 4,612 \\
\bottomrule
\end{tabular}
\end{center}
\caption{\label{tab_lbl_stat} Dataset label distribution, ranked by frequency. }
\end{table}

Table~\ref{dataset_comp} shows that \textsc{KinNews} is divided in 17,014 articles for training and 4,254 articles for the test set, with an average of 302.9 words per article and approximately 370K unique words (when lowercased) for the raw version, and average of 246.5 words per article and approximately 300K unique words (when lowercased) for the cleaned version. \textsc{KirNews} which is quite small compared to \textsc{KinNews}, is divided in 3,690 articles for train set and 922 articles for test set, with an average of 264.5 words per article and approximately 86K unique words (when lowercased) for the raw version, and average of 210.2 words per article and approximately 63K unique words (when lowercased) for the cleaned version. The comparison with other high-resource news classification datasets for English, namely \textit{AG News}\footnote{\url{http://groups.di.unipi.it/~gulli/AG_corpus_of_news_articles.html}} and \textit{Reuters} (ApteMod)~\cite{lewis1997reuters} and low-resource news datasets (Setswana and Sepedi \cite{marivate-etal-2020-investigating}) shows that our datasets are composed of the longest news articles and have the largest vocabulary size, which indicates their suitability for training large deep learning models and NLP studies in general.

\begin{table}[ht]
\begin{center}
\begin{tabular}{llcrrrcr}
\toprule
\bf Resources & \bf Dataset& \bf Classes & \Longstack{\bf Train\\\bf Samples} & \Longstack{\bf Test\\\bf Samples} & \bf Total & \Longstack{\bf Average \\\bf Words/Article} & \Longstack{\bf Unique\\\bf Words} \\ \midrule
{\multirow{2}{*}{\Longstack{High-\\resource}}} & AG's News & 4 & 120,000 & 7,600 & 127,600 & 37.8 & 188,110 \\
&\Longstack{Reuters} & 90 & 7,769 & 3,019 & 10,788 & - & - \\ \midrule
{\multirow{2}{*}{\Longstack{Low-\\resource}}} & \Longstack{Setswana}& 10 & - & - & 219 & - & 1,561 \\
&\Longstack{Sepedi} & 10 & - & - & 491 & - & 3,018 \\ \midrule
{\multirow{4}{*}{\Longstack{Low-\\resource \\(Ours)}}} & \Longstack{\textsc{KirNews}}& 12 & 3,690 & 922 & 4,612 & 264.5 & 86,023 \\
&\Longstack{\textsc{KirNews}*} & 12 & 3,690 & 922 & 4,612 & 210.2 & 63,143 \\ 
&\Longstack{\textsc{KinNews}} & 14 & 17,014 & 4,254 & 21,268 & 302.9 & 370,036 \\ 
&\Longstack{\textsc{KinNews}*} & 14 & 17,014 & 4,254 & 21,268 & 246.5 & 301,016 \\
\bottomrule
\end{tabular}
\end{center}
\caption{\label{dataset_comp} Comparison with other news classification datasets. \textsc{KirNews}* and \textsc{KinNews}* denote cleaned versions of \textsc{KirNews} and \textsc{KinNews}, respectively.  }
\end{table}

An in-depth evaluation of the similarity between Kinyarwanda and Kirundi using the created datasets shows that they share 27,489 words of the vocabularies which stands for 32\% of all unique words from \textsc{KirNews}, using the raw version datasets and 22,841 vocabularies which stands for 36.2\% of all unique words from \textsc{KirNews}*, using the cleaned versions of the datasets.

\section{Experiments}\label{sec:experiments}
\subsection{Word Embedding Training}
Many African low-resource languages, including Kinyarwanda and Kirundi, do not enjoy the success of recent word embeddings available as pre-trained models such as GloVe \cite{pennington2014glove}, BERT \cite{devlin2018bert}, XLNet \cite{yang2019xlnet}, or Fasttext \cite{grave2018learning} because these models were trained on higher-resource languages exclusively. Recent text classification approaches for low-resource languages rely on transfer learning approach that uses the features of resource-rich languages learned by pre-trained word embeddings to train low resource models. However, this technique might not be effective enough or even not be applicable when there is no parallel corpus of that resource-rich and resource-poor languages.

Since our datasets contain a reasonable amount of sentences, the features to train our neural-network-based models are obtained by training Word2Vec embeddings \cite{mikolov2013efficient} from scratch. Word2Vec is trained using the \texttt{gensim} framework\footnote{\url{https://radimrehurek.com/gensim/models/word2vec.html}} with a window size of 5, ignoring all words with total frequency lower than 5, removing stopwords, special characters and URLs, and using skip-gram training algorithm with hierarchical softmax. We train two versions with different dimensions on each language, one with 50 dimensions (\texttt{W2V-Kin-50}) and other with 100 dimensions (\texttt{W2V-Kin-100}) for Kinyarwanda, and \texttt{W2V-Kir-50} and \texttt{W2V-Kir-100} for Kirundi. 

\subsection{Text Classification Task}

\paragraph{Monolingual}
For a monolingual approach, we train and evaluate our baseline models using \textsc{KinNews} for Kinyarwanda and \textsc{KirNews} for Kirundi separately. This means that we are using exclusively the data available for each task, ignoring the similarity of both languages.

\paragraph{Cross-lingual}
Cross-lingual transfer has been leveraged for many low-resource applications \cite{agic2015if,buys2016cross,adams2017cross,fang2017model,cotterell2017low}. Most commonly, these approaches rely on machine translation and word alignments between resource-rich and low-resource languages. In this paper, however, we follow a simpler approach that exploits the fact that both languages are mutually intelligible, and does not require parallel or aligned resources. We train the baseline models using \textsc{KinNews} and embeddings learned from Kinyarwanda, and test them on \textsc{KirNews}. This simulates the scenario if we did not have any training data for Kirundi. Alternatively, we train and test embedding-based models on \textsc{KirNews} using the Kinyarwanda embeddings. This models a scenario where embeddings in a higher-resourced related language are available, and a small training set in the target language. We only investigate the transfer from Kinyarwanda to Kirundi and not the reverse, since our Kinyarwanda data is much larger than Kirundi data.\footnote{
We remove \textit{tourism} and \textit{fashion} related samples from \textsc{KinNews} to get a compatible training set for the \textsc{KirNews} test set which does not contain articles from these two categories.}

\subsection{Baseline Models}
We perform benchmark experiments on the datasets using several different classic and neural approaches. In all experiments, we use the pre-processed (cleaned) version datasets, because the raw versions contain too much noise. The training set and validation set are split with a ratio of 9:1.

\subsubsection{Classic Models}
For all classic machine learning approaches, we use Term Frequency Inverse Document Frequency (TF-IDF) to get the values of unigram input features. We define the maximum number of features to be used depending on different train set and method. All of the below models are implemented with the help of the \texttt{scikit-learn} framework and use its default hyperparameters:
\begin{itemize}
\item\textbf{ Multinomial Naive Bayes (MNB)} 

\item\textbf{ Logistic Regression (LR)}

\item \textbf{Support Vector Machine (SVM) with SGD}
\end{itemize}
\subsubsection{Neural Models}
For neural models, we use the pre-trained embeddings as input and fine-tune them on the task (except for character-based models). This is to mimic approaching an arbitrary NLP task with little training data but with available word embeddings. The following neural models are implemented:
\begin{itemize}
\item \textbf{Character-level Convolutional Neural Networks (Char-CNN):} We use a small size Char-CNN model for text classification as proposed in \cite{zhang2015character} with default hyperparameters, except that we removed the letters \textit{'q'} and \textit{'x'} from the alphabet list which are not in both Kinyarwanda and Kirundi languages. Thus, the alphabet used in our model consists of 68 characters instead of 70 characters from the original paper. The input feature length is also changed from 1,014 to 1,500 to capture most of the texts of interest, since our datasets have relatively long news articles. The special properties of Char-CNN that makes it a good choice for low-resource languages are that (1) it does not require any data preprocessing nor (2) the use of word embeddings which makes it more effective when processing very noisy data.
\item \textbf{Convolutional Neural Network (CNN):} We use the CNN for sentence classification model proposed in \cite{kim2014convolutional} with default hyperparameters, except that we change the original feature maps of 100 to 150 and min-batch size of 50 to 32. The model is trained on two Word2Vec embeddings with different dimensions of 50 and 100, and using different epochs and number of features based on different train sets and embedding dimensions.
\item \textbf{Bidirectional Gated Recurrent Unit (BiGRU):} We design a model that consists of 2-layer bidirectional GRU \cite{cho2014learning} followed by a softmax linear layer. It uses the dropout of 0.5 and batch size of 32. The dimension of hidden layers were set to either 256 or 128, it is trained on two Word2Vec embeddings with different dimensions of 50 and 100 similar to CNN, and different epochs and number of features were used according to different train sets and embedding dimensions similar to the previous models.
\end{itemize}
\subsection{Results}

\subsubsection{Monolingual Text Classification}
The experimental results for monolingual text classification are shown in Tables~\ref{tab_kin_res} and~\ref{tab_kir_res}, respectively. In each table the benchmark results of the classic TFIDF-based models and the neural embedding-based models are grouped separately, and we highlight the result of the best model in each group.

\begin{table}[ht]
\begin{center}
\begin{tabular}{llccc}
\toprule
\bf Model & \bf Embeddings & \bf Features ($\times$1000) & \bf Epochs & \bf Accuracy (\%) \\ \midrule
MNB & TF-IDF & 6 & - & 82.70 \\
LR & TF-IDF & 20 & - & 87.14 \\
SVM & TF-IDF & 90 & - & \bf 88.53 \\ \midrule
Char-CNN & - & - & 20 & 71.70 \\ 
{\multirow{2}{*}{CNN}} & \texttt{W2V-Kin-50} & 15 & 8 & 87.55 \\
& \texttt{W2V-Kin-100} & 40 & 4 & 87.54 \\ 
{\multirow{2}{*}{BiGRU}} & \texttt{W2V-Kin-50}$^\ast$ & 10 & 10 & \bf 88.65 \\
& \texttt{W2V-Kin-100} & 10 & 6 & 88.29 \\ \bottomrule
\end{tabular}
\end{center}
\caption{\label{tab_kin_res} Benchmark results on the \textsc{KinNews} dataset. The $^\ast$ on \texttt{W2V-Kin-50} for BiGRU denotes that the dimension of hidden layers was set to 128 instead of 256. }
\end{table}

\begin{table}[ht]
\begin{center}
\begin{tabular}{llccc}
\toprule
\bf Model & \bf Embeddings & \bf Features ($\times$1000) & \bf Epochs & \bf Accuracy (\%) \\ \midrule
MNB & TF-IDF & 4 & - & 82.67 \\
LR & TF-IDF & 6 & - & 86.13 \\
SVM & TF-IDF & 63 & - & \bf 90.14 \\ \midrule
Char-CNN & - & - & 20 & 69.23 \\ 
{\multirow{2}{*}{CNN}} & \texttt{W2V-Kir-50} & 40 & 8 & 85.75 \\
& \texttt{W2V-Kir-100} & 35 & 4 & \bf 88.01 \\ 
{\multirow{2}{*}{BiGRU}} & \texttt{W2V-Kir-50} & 10 & 12 & 85.86 \\
& \texttt{W2V-Kir-100} & 10 & 6 & 86.61 \\ \bottomrule
\end{tabular}
\end{center}
\caption{\label{tab_kir_res} Benchmark results on the \textsc{KirNews} dataset. }
\end{table}

As shown in Table~\ref{tab_kin_res} and \ref{tab_kir_res}, in the group of classic TFIDF-based models, SVM yields the best accuracy on both datasets compared to LR and MNB. It has high predictive power thanks to the hyperplane which can avoid the overfitting and separates the classes in very effective way. Another good property of SVM is that by using its supporting vectors, it can use relatively small amount of data to get a good prediction, which makes it to perform well on both \textsc{KinNews} and \textsc{KirNews}. In this group, MNB performs the worst on both datasets, however, it was able to give relatively good results by requiring fewer features compared to other methods. 

In the group of neural embedding-based models, the performance is based on the type of data, where BiGRU perform the best on \textsc{KinNews} which is larger than \textsc{KirNews} dataset, while CNN perform the best on \textsc{KirNews}. A possible reason might be that BiGRU needs a larger amount of data to perform better than the CNN. The Char-CNN performs the worst on both datasets, likely because of the limited computation power of the compute resources used in the experiments. Because of this we had to limit the input feature length to 1500 while the average length of the news article in each dataset was far greater than that. Thus it is an open challenge to further work that can be able to use higher length of input features to achieve better results. It might also be interpreted as a pointer towards the general ``pre-train and fine-tune'' regime, which places classification models in an initial representation space that reflects word relations in the input.

\begin{table}[ht]
\begin{center}
\resizebox{\columnwidth}{!}{%
\begin{tabular}{llllccc}
\toprule
\bf Train set & \bf Test Set & \bf Model & \bf Embeddings & \bf Features ($\times$1000) & \bf Epochs & \bf Accuracy (\%) \\ \midrule
{\multirow{3}{*}{\textsc{KinNews}}} & {\multirow{3}{*}{\textsc{KirNews}}} & MNB & TF-IDF & 6 & - & \bf 73.46 \\
&&LR & TF-IDF & 6 & - & 68.26 \\
&&SVM & TF-IDF & 35 & - &  72.70 \\ \midrule
{\multirow{5}{*}{\textsc{KinNews}}} & {\multirow{5}{*}{\textsc{KirNews}}} & Char-CNN & - & - & 6 & 49.60 \\ 
&&{\multirow{2}{*}{CNN}} & \texttt{W2V-Kin-50} & 30 & 5 & 60.64 \\
&&& \texttt{W2V-Kin-100} & 20 & 4 & 61.72\\ 
&&{\multirow{2}{*}{BiGRU}} & \texttt{W2V-Kin-50} & 10 & 7 & \bf 67.54 \\
&&& \texttt{W2V-Kin-100}$^\ast$ & 10 & 7 & 65.06 \\ \midrule \midrule
{\multirow{4}{*}{\textsc{KirNews}}} & {\multirow{4}{*}{\textsc{KirNews}}} & {\multirow{2}{*}{CNN}} & \texttt{W2V-Kin-50} & 40 & 12 & 85.75 \\ 
&&& \texttt{W2V-Kin-100} & 35 & 8 & \bf 88.01 \\ 
&&{\multirow{2}{*}{BiGRU}} & \texttt{W2V-Kin-50} & 10 & 12 & 83.38 \\
&&& \texttt{W2V-Kin-100} & 10 & 10 & 86.61 \\ \bottomrule
\end{tabular}
}
\end{center}
\caption{\label{tab_cross_res} Benchmark results for cross-lingual approaches.The $^\ast$ on \texttt{W2V-Kin-100} for BiGRU denotes that the dimension of hidden layers was set to 128 instead of 256. }
\end{table}

\subsubsection{Cross-lingual Text Classification}
The results on cross-lingual approaches in Table~\ref{tab_cross_res} show that in the group of machine learning models, MNB surprisingly performs relatively better compared to SVM and LR. The reason might be that MNB is a generative model while the rest are discriminative models. 

Similar to the monolingual experiments, BiGRU performs the best when trained on \textsc{KinNews} while CNN performs the best when trained on \textsc{KirNews}. Interestingly, the Char-CNN suffers much more from the transfer, which illustrates that the embeddings reflect the similarities of both languages on an abstract level that allow transfer much better than low-level features trained from scratch.

When trained neural models on \textsc{KinNews} and testing them on \textsc{KirNews}, they do not give satisfactory results compared to when trained and tested on only \textsc{KirNews}. This is not surprising, since they were both trained on Kinyarwanda word embeddings from the same domain which includes the word vectors of many similar words from Kirundi. Nevertheless, this shows that with pretrained Kinyarwanda word embeddings, which are easier to obtain in high quality since the data retrieval for Kinyarwanda is much easier, can be effectively used in training Kirundi text classification models in a zero-shot scenario without any labeled or unlabeled data for Kirundi.

Based on the performed experiments, the results for our examplary languages Kinywarwanda and Kirundi show that the text classification based on cross-lingual transfer between mutually intelligible low-resource languages is possible,  without creating any words alignments or any parallel translation dataset between those languages. What is merely required is to get sufficient data of one language to train the word embeddings.

Analysing the classification errors of highest-scoring cross-lingual models, we find that the MNB model characteristically places many articles about ``education'' in the category of ``politics'', while the neural models are more accurate in this distinction. All models tend to confuse ``relationship'' articles with the ``politics'' category, which might be due to an overlap in common vocabulary focused on interactions of people. The most accurate classification is generally obtained for sports articles. Complete confusion matrices are displayed in Appendix~\ref{appendix:confusion-matrices}.

\section{Conclusion and Future Work}\label{sec:conclusion}
In this paper, we built the first news text classification benchmark for Kinyarwanda and Kirundi, two low-resource Bantu languages. We described the data collection process, provided guidelines for data cleaning, and evaluated classic text classification models as initial baselines. We found fairly strong cross-lingual generation of embedding models trained on the resource-richer Kinyarwanda to the lower-resource Kirundi due to their mutual intelligibility. This gives hope for future studies of languages from the Rwanda-Rundi language family, which have not been studied in NLP at all, and would otherwise classify as ``Left-Behind'' according to~\cite{joshi-etal-2020-state}, or analogously of other extremely low-resource languages with a slightly higher-resource ``sibling''.

Future studies on the new dataset will investigate (1) contextualized embeddings, e.g. BERT, (2) sub-word modeling. Furthermore, the dataset may get enriched with other linguistic annotations, such as named entities, and serve as resource for other NLP tasks than text classification.

\section*{Acknowledgements}
 This work was partially supported by the National Key R\&D Program of China under Grant 2018AAA0100202, and the National Science Foundation of China under Grants 61976043. We thank reviewers for their helpful and insightful comments. We also give special thanks to all Masakhane members for the encouragement and guidance and Ndimubandi Samuel de Jesus for the useful discussion on Kirundi language.  

\bibliographystyle{coling}
\bibliography{coling2020}

\newpage
\begin{appendices}
\section{Annotation Rules}
\label{appendix:table}
\begin{longtable}{ p{.10\textwidth}  p{.10\textwidth} p{.10\textwidth}  p{.22\textwidth}  p{.10\textwidth}  p{.23\textwidth} }
\toprule
\multicolumn{2}{c}{\bf Shared Labels}& \multicolumn{2}{c}{\bf \textsc{KinNews}} &\multicolumn{2}{c}{\bf \textsc{KirNews}} \\ \midrule
 Label & En\_Label & Kin\_Label & Initial\_Category & Kir\_Label & Initial\_Category\\ \hline
 {\multirow{7}{*}{1}} & {\multirow{7}{*}{politics}} & {\multirow{7}{*}{politiki}} & politiki (politics) & {\multirow{7}{*}{poritike}} & politike (politics) \\
&&&imiyoborere (governance) & & imibano (int'l relations) \\
&&&diyasipora (diaspora) & &diyasipora (diaspora)  \\
&&&amatora (elections) & & imigenderanire(relations) \\
&&&ubutabera (justice) & & ubutungane (justice) \\
&&&umutekano (security) & & umutekano (security) \\
&&&kwibuka (genocide memorial) & & impunzi (refugee news) \\ \midrule
{\multirow{9}{*}{2}} & {\multirow{9}{*}{sport}} & {\multirow{9}{*}{imikino}} & imikino (sport) & {\multirow{9}{*}{inkino}} & inkino (sport) \\
&&&siporo (sport) & &  \\
&&&football & &  \\
&&&basketball & &  \\
&&&volleyball & & \\
&&&amagare (biking) & & \\
&&&handball & & \\
&&&ngororangingo (stretching) & & \\
&&&karate & & \\ \midrule
{\multirow{9}{*}{3}} & {\multirow{9}{*}{economy}} & {\multirow{9}{*}{ubukungu}} & ubukungu(economy) & {\multirow{9}{*}{ubutunzi}} & ubutunzi (economy) \\
&&&ubucuruzi (business) & &ubudandaji (business) \\
&&&ubuhinzi (agriculture) & & uburimyi(agriculture) \\
&&&iterambere (development)& & iterambere (development) \\
&&&ubworozi (farming) & & ubworozi (farming) \\
&&&ubwiteganyirize (insurance) & & indanga (business) \\
&&&imigabane (stock market) & &  \\
&&&ivunjisha (exchange) & & \\ 
&&&ishoramari (investment) & & \\ \midrule
{\multirow{2}{*}{4}} & {\multirow{2}{*}{health}} & {\multirow{2}{*}{ubuzima}} & ubuzima (health) & {\multirow{2}{*}{amagara}} & amagara (health) \\
&&&indwara (disease) & &ubuzima (health) \\ \midrule
{\multirow{8}{*}{5}} & {\multirow{8}{*}{\Longstack{enterta-\\inment}}} & {\multirow{8}{*}{\Longstack{imyida-\\gaduro}}} & \Longstack{imyidagaduro (\\entertainment)} & {\multirow{8}{*}{\Longstack{kwida-\\gadura}}} & \Longstack{kwidagadura (\\entertainment)} \\
&&&showbiz & &  \\
&&&indirimbo (songs) & &  \\
&&&muzika (music) & & \\
&&&sinema (cinema) & & \\
&&&ibirori (party) & & \\
&&&urwenya (comedy) & & \\ \midrule
{\multirow{2}{*}{6}} & {\multirow{2}{*}{history}} & {\multirow{2}{*}{amateka}} & amateka (history) & {\multirow{2}{*}{akahise}} & akahise (history) \\
&&&mu mateka (in history) & & \\ \midrule
7& \Longstack{techn-\\ology} &\Longstack{ikorana-\\buhanga}& \Longstack{ikoranabuhanga (\\technology)}& ubuhinga &\Longstack{ubuhinga (technology)} \\ \midrule
8& tourism &\Longstack{ubuke-\\rarugendo}& \Longstack{ubukerarugendo (tourism)}& - &- \\ \midrule
9& culture & umuco & umuco (culture) & akaranga & imicokama (culture) \\ \midrule
10& fashion & imideli & imideli (fashion) & - & - \\ \midrule
{\multirow{2}{*}{11}} & {\multirow{2}{*}{religion}} & {\multirow{2}{*}{\Longstack{iyobo-\\kamana}}} &  \Longstack{iyobokamana (religion)} & {\multirow{2}{*}{ukwemera}} & ukwemera (religion) \\
&&& & &ivy'Imana (religion) \\ \midrule
12& \Longstack{envir-\\onment} & ibidukikije & \Longstack{ibidukikije (environment)} & ibidukikije & \Longstack{ibidukikije (environment)} \\ \midrule
{\multirow{3}{*}{13}} & {\multirow{3}{*}{education}} & {\multirow{3}{*}{uburezi}} & uburezi (education) & {\multirow{3}{*}{indero}} & indero (education) \\
&&&ubumenyi (knowledge) & & inyigisho (lessons) \\ 
&&&mu mashuli (schools) & & \\ \midrule
{\multirow{3}{*}{14}} & {\multirow{3}{*}{\Longstack{relat-\\ionship}}} & {\multirow{3}{*}{urukundo}} & urukundo (relationship) & {\multirow{3}{*}{urukundo}} &  urukundo (relationship) \\
&&&gushaka umukunzi (dating) & & \\ 
&&&ubukwe no gusaba (weddings) & & \\ 
\bottomrule

\caption{\label{tab1} Annotation category combinations based rules. }
\end{longtable}


\section{Confusion Matrices of our best Cross-Lingual Models}
\label{appendix:confusion-matrices}
\begin{figure}[ht]
\centering
\includegraphics[width=0.50\columnwidth]{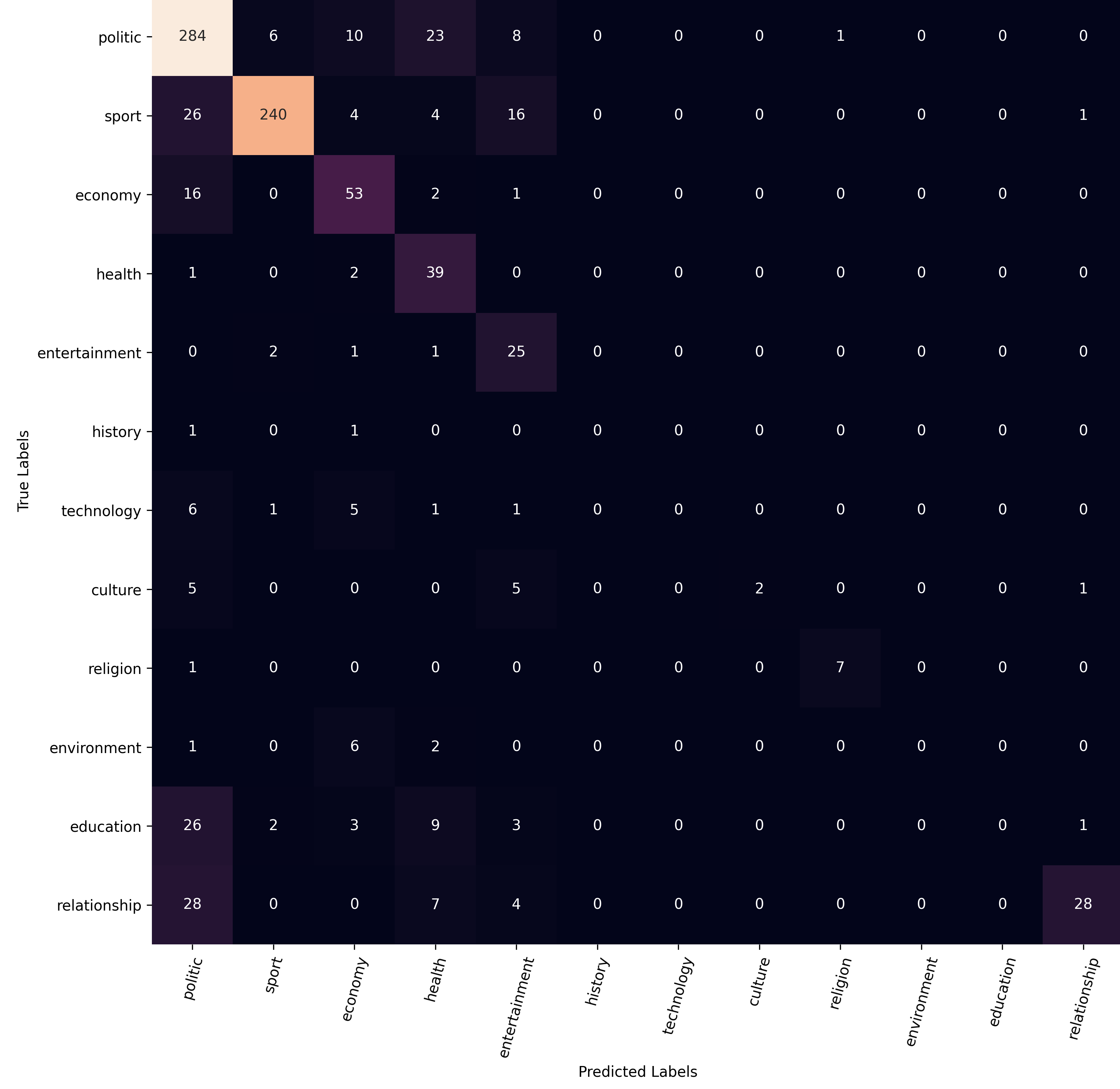}
\caption{\label{cm-cross-bst-models-1} Confusion matrices of TF-IDF+MNB.} 
\end{figure}

\begin{figure}[ht]
\centering
\includegraphics[width=0.50\columnwidth]{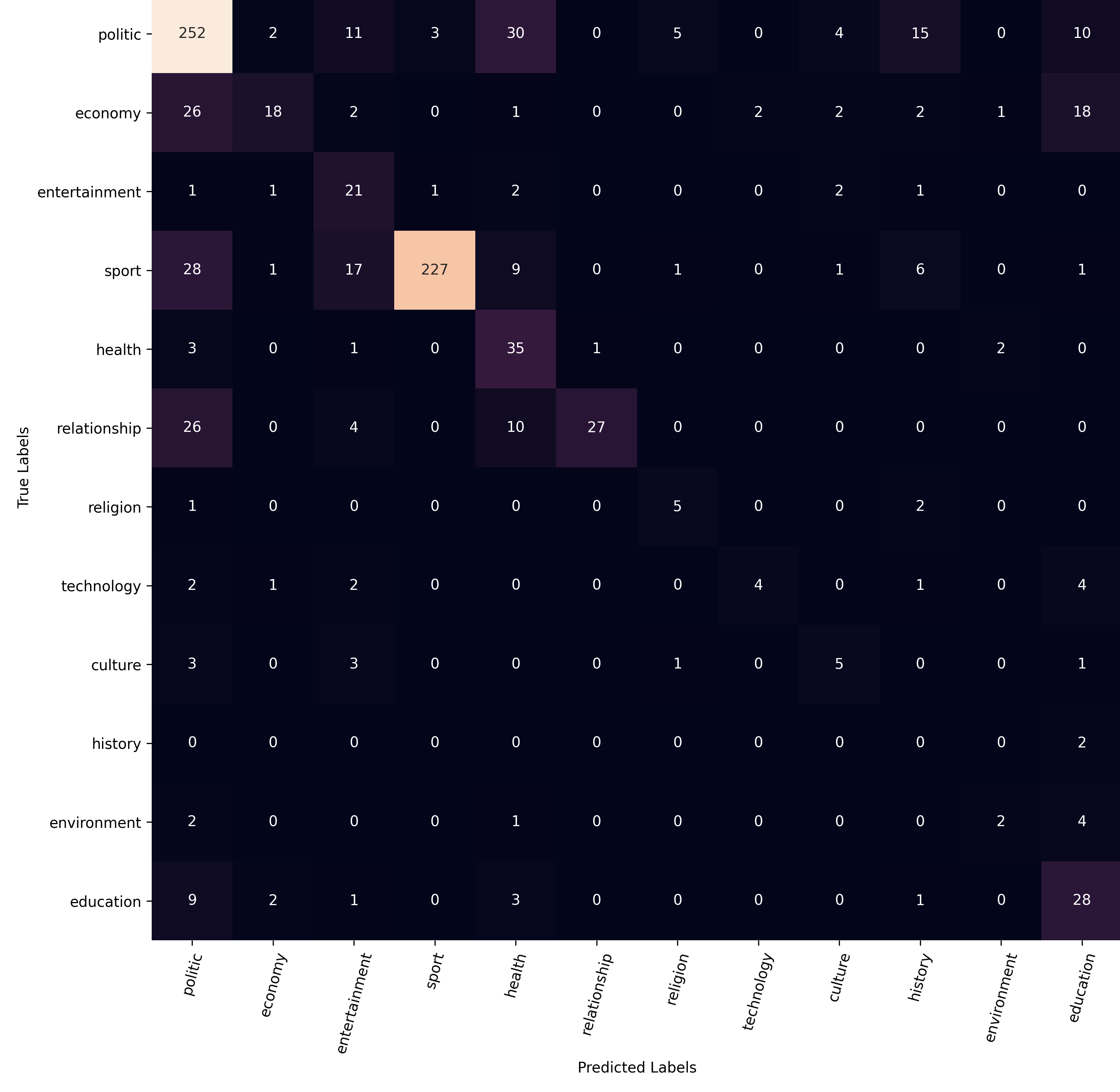}
\caption{\label{cm-cross-bst-models-2} Confusion matrices of \texttt{W2V-Kin-50}+BiGRU.} 
\end{figure}

\begin{figure}[ht]
\centering
\includegraphics[width=0.50\columnwidth]{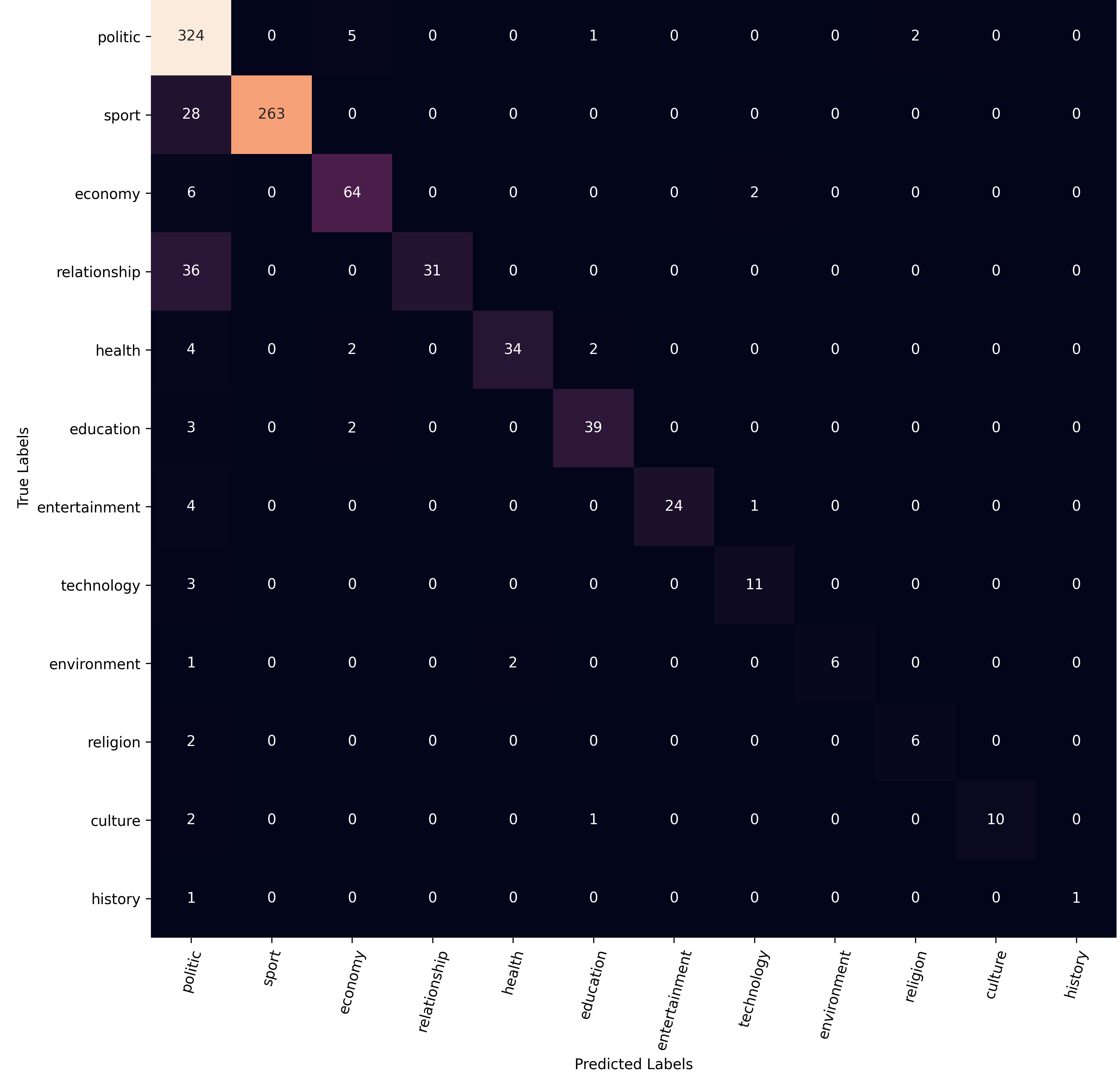}
\caption{\label{cm-cross-bst-models-3} Confusion matrices of \texttt{W2V-Kin-50}+CNN.} 
\end{figure}

\end{appendices}
\end{document}